\documentclass[final,3p,times,twocolumn]{elsarticle}

\usepackage{amssymb}
\usepackage{arydshln} 
\usepackage[table]{xcolor}
\usepackage{colortbl}
\usepackage{graphicx}
\usepackage{amsmath}
\usepackage{amssymb}
\usepackage{booktabs}
\usepackage{enumitem}
\usepackage{multirow}
\usepackage{graphicx}
\usepackage{float} 

\usepackage{makecell}

\usepackage{graphicx}
\definecolor{Gray}{gray}{0.9}

\usepackage{xcolor}
\usepackage{caption}
\usepackage{currfile}

\newcommand{\re}[1]{\textcolor{black}{#1}}
\newcommand{\rcap}{\captionsetup{labelfont={color=black}, textfont={color=black}}}

\newcommand{\assetpath}{}

\begin{document}

\begin{frontmatter}


\title{StereoDiffuer: Diffusion-based Progressive Geometry Modeling \\with Saliency Attention Perception for Stereo Matching}

\author[label1,label2]{Bohan Li} 

\affiliation[label1]{organization={Shanghai Jiaotong University},
            addressline={No.800 Dongchuan Road}, 
            city={Shanghai},
            postcode={200240}, 
            country={China}}

\affiliation[label2]{organization={Ningbo Institute of Digital Twin, Eastern Institute of Technology},
            addressline={No. 568, Tongxin Road}, 
            city={Ningbo},
            postcode={315000}, 
            country={China}}


\begin{abstract}
With the advance of deep neural networks, the quality of disparity maps obtained through stereo matching has steadily improved. However, existing stereo matching methods still struggle to preserve fine-grained geometric details, resulting in blurred edges and over-smoothed predictions in challenging regions. To address these limitations, we propose StereoDiffuer, an iterative diffusion-based stereo matching framework that explicitly models geometric details and progressively refines disparity estimates. The framework incorporates a Saliency Attention Perception (SAP) module to extract salient geometric cues, including object boundaries, thin structures, and sharp edges. Confidence-guided SAP features are combined with the initial disparity estimate to condition an iterative denoising diffusion process, which corrects residual disparity errors and restores geometric details suppressed during cost-volume regularization and upsampling. \re{Experimental results on the Scene Flow and KITTI benchmarks demonstrate the effectiveness of the proposed framework and its competitive performance relative to the compared stereo matching methods.}
\end{abstract}




\end{frontmatter}

\section{Introduction}
Establishing correspondences in two-view images, commonly known as binocular stereo matching, plays a pivotal role in 3D visual perception. It constitutes a cornerstone for numerous 3D vision applications, including autonomous driving, robot navigation, augmented reality and industrial manufacturing~\cite{lin2017optimizing,nguyen2019wide,li2019depth,zhang2021dense,yuan2021efficient,wang2023multi,lei2017depth,okae2021robust}. Recent years have witnessed profound advancements in this domain, with deep neural networks showcasing exceptional performance in addressing highly complex real-world scenarios~\cite{fan2023accurate,wen2022feature,cheng2022region,zbontar2015computing, guney2015displets, kendall2017end, chang2018pyramid, guo2019group,li2022improved,li2024time}. Prevailing state-of-the-art network architectures continue to confront challenges in handling ambiguities in correspondence matching, particularly in challenging scenarios involving textureless regions, repetitive patterns, slender structures, occlusions, and reflective surfaces. Such difficulties lead to the prevalence of inaccurate disparity predictions~\cite{lu2018sparse, wang2021scv,zhang2019dispsegnet,peng2022rethinking,xu2023iterative,li2024time}.

Existing methods commonly rely on regularization to cope with matching ambiguities in stereo matching. Early methods~\cite{vzbontar2016stereo,shaked2017improved,mayer2016large} apply feature correlation in conjunction with 2D convolutions to encode and regularize matching cost volumes, from which the optimal disparity values are selected. Subsequent approaches~\cite{kendall2017end, chang2018pyramid, okae2021robust} demonstrate that feature concatenation and 3D-CNN-based cost-volume regularization can improve matching quality at higher computational cost. To improve efficiency, recent techniques~\cite{liang2019stereo, wu2019semantic, yang2019hierarchical, gu2020cascade, cfnet, wang2021patchmatchnet} adopt coarse-to-fine (CTF) stereo matching frameworks that progressively fuse information from multi-scale cost volumes to reduce computational cost.

Despite these achievements in boosting matching precision and computational efficiency, the issue of blurred edges and over-smoothed boundaries persists in the estimated disparity maps~\cite{zhang2019ga, xu2020aanet, yu2018deep}.
These artifacts stem from insufficient extraction and preservation of fine-grained geometric details.
To mitigate these challenges, another line of research has introduced GRU-based architectures~\cite{Lahav2021raftstereo,jiankun2022crestereo,cai2022riav,xu2023iterative} to facilitate dependable disparity estimation through iterative
refinement. Nevertheless, these approaches typically use low-level contextual features to refine the initial disparity without explicitly modeling fine-grained geometric details~\cite{Lahav2021raftstereo,cai2022riav,jiankun2022crestereo}. Moreover, ConvGRU-based refinement may accumulate errors over recurrent updates~\cite{li2017diffusion, mao2022review}.\looseness=-1

\begin{figure}[t]
\centering
\includegraphics[width=0.99\linewidth]{\assetpath 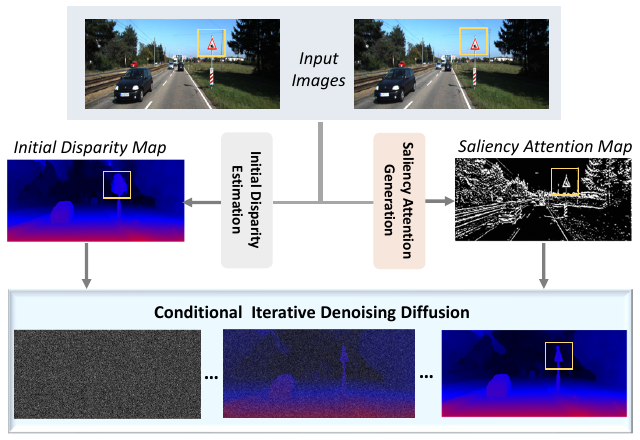}
\caption{Overall pipeline of the proposed StereoDiffuer framework. 
Given a pair of stereo images, we first estimate an initial disparity map and extract a saliency attention map that highlights high-frequency geometric cues, such as object boundaries, thin structures, and sharp edges. 
The initial disparity map provides a coarse geometric prior, while the saliency attention map supplies detail-aware guidance. 
Both are used to construct the condition for the iterative denoising diffusion module, which progressively corrects disparity errors and refines over-smoothed boundary regions. 
The highlighted road sign illustrates how the proposed refinement improves geometric details in challenging areas.\looseness=-1}
\label{teaser}
\end{figure}

Based on the above analysis, we propose to extract distinctive geometric information from the input stereo images to supplement the initial ``imperfect'' disparity.
Departing from the use of GRU mechanisms as adopted in prior research~\cite{Lahav2021raftstereo,cai2022riav,jiankun2022crestereo}, we resort to harnessing generative denoising diffusion models to progressively refine the estimation results based on a Markov chain~\cite{saharia2022image,muller2022diffrf,li2024time}. As a recently emerged generative modeling paradigm, diffusion models construct intermediate target distribution at each step of the Markov chain, which effectively decomposes the challenging task of learning intricate empirical distributions into a series of multi-step distribution transitions~\cite{muller2022diffrf,li2024time}. Concerning the complexity of 3D scene perception, this sequential modeling strategy is particularly applicable to our stereo matching task. It enables us to convert the complex disparity prediction into a progressive depth distribution modeling process.

To this end, we introduce a Saliency Attention Perception (SAP) module to capture geometric knowledge and fine-grained details, which are integrated into conditional iterative denoising diffusion to progressively optimize the initial ``imperfect'' disparity maps.
The Saliency Attention Perception module establishes attention maps to explicitly indicate the high-frequency details of edges, sharp object boundaries and other critical visual saliency features within the input images.
The diffusion model is conditioned on the initial disparity map and the saliency attention map to progressively model reliable geometric distributions.
The two sources are complementary: the initial disparity map provides a coarse geometric prior, while the saliency attention map identifies fine-grained geometric details. \re{As illustrated in Figure~\ref{teaser}, the proposed refinement visibly reduces matching errors and boundary over-smoothing in the displayed road-sign example.}

The proposed StereoDiffuer is designed around a task-specific observation in stereo matching: cost-volume aggregation and disparity upsampling can provide reliable global geometry, but they often suppress high-frequency structures around object boundaries, thin objects, and sharp edges. 
To recover these missing geometric details, we introduce SAP to extract saliency-aware boundary and structure cues from image features, and use the cost-volume confidence to adaptively select reliable saliency information. 
The selected saliency features are then combined with the initial disparity map to construct a collaborative condition for the denoising diffusion module. Under this condition, the diffusion process progressively refines the initial disparity by correcting residual errors in geometrically challenging regions, rather than serving as a generic post-processing step.\looseness=-1
The main contributions of our work are summarized as follows:
\begin{itemize}

\item We introduce a deep saliency attention mechanism to explicitly extract geometric knowledge and distinguish high-frequency details to tackle the over-smoothness in stereo matching.

\item We propose to condition the iterative denoising diffusion model with the complementary initial disparity maps and saliency attention maps for reliable continuous disparity refinement.

\item Experimental results on Scene Flow~\cite{mayer2016large} and KITTI~\cite{geiger2012we} benchmarks illustrate that our proposed method is effective and provides accurate and geometrically consistent disparity maps.

\end{itemize}

\section{Related Work}
\subsection{Learning-based Binocular Stereo Matching}

Recent developments in stereo matching focus on employing deep neural networks for correspondence computation~\cite{mayer2016large,vzbontar2016stereo,pang2017cascade,xu2022attention,xu2023iterative,li2024bridging,watson2021temporal,long2021multi,cai2023riav}. Zbontar et al.~\cite{zbontar2015computing} first propose a Siamese network to predict the similarity of two image patches. Mayer et al.~\cite{mayer2016large} introduce a 2D convolution-based architecture in an end-to-end fashion without post-processing. CRL \cite{pang2017cascade} constructs a two-stage network with cascade residual learning to further improve the performance. However, these approaches have difficulties in producing a reliable cost volume for accurate disparity regression.
This challenge has motivated several techniques to improve the cost volume construction. GCNet~\cite{kendall2017end} concatenates corresponding features from two input images, and directly feeds them into a 3D regularization module to aggregate contextual information.
In PSMNet~\cite{chang2018pyramid}, an hourglass deep network is introduced, which further employs 3D convolutions in cost aggregation. Yu et al.~\cite{yu2018deep} present a two-stream network that incorporates global view guidance. A hybrid approach~\cite{guo2019group} combines explicit correlation and concatenation in cost volume construction. Zhang et al.~\cite{zhang2019ga} leverage the concepts of guided cost filter and semi-global matching to design a two-layer guided aggregation network for cost volume regularization. AANet~\cite{xu2020aanet} introduces intra-scale and cross-scale aggregation modules to replace the computationally expensive 3D convolutions.
RSSM~\cite{cheng2022region} proposes a two-stage network to construct and process of cost volumes from different regions separately.

A number of approaches have adopted the CTF framework to mitigate the computational burden in 3D aggregation, while preserving rich global contextual information~\cite{deng2021detail,dovesi2020real,dai2021adaptive,cheng2020deep,gu2020cascade,li2022practical,peng2022rethinking,li2023nr}. MCV-MFC~\cite{liang2019stereo} uses a multi-level cost volume scheme to improve the robustness of the network and refine the disparity in an efficient manner. SSPCV-Net~\cite{wu2019semantic} constructs heterogeneous and multi-scale cost volumes combined with semantic features, and introduces a 3D multi-cost aggregation module. CFNet~\cite{cfnet} fuses multiple cost volumes and gradually refines the disparities across different levels. Despite its effectiveness, the CTF framework often suffers from edge blurriness, which leads to incorrect disparity predictions.
Another line of research refines disparity estimates using supplementary information and attention maps. EdgeStereo~\cite{song2020edgestereo} uses edge cues, including edge-feature embedding and an edge-aware smoothness loss, to guide disparity optimization. SegStereo~\cite{yang2018segstereo} introduces semantic information for disparity refinement, while Zhang et al.~\cite{zhang2019dispsegnet} concatenate semantic cues with the initial disparity in an unsupervised framework. Other methods compute and fuse multi-scale disparity maps hierarchically~\cite{yang2019hierarchical,wang2021patchmatchnet,xu2020aanet}, or introduce disparity-upsampling schemes for refinement in a coarse-to-fine stereo framework~\cite{nie2019multi,xu2021bilateral,zhang2021learning}. In CNN-based refinement, effectiveness depends on the quality and saliency of the guidance information~\cite{sang2019multi,yu2018deep,song2020edgestereo,xu2022attention}. In contrast, we explicitly model reliable, fine-grained geometric cues with a deep saliency attention mechanism for disparity refinement.\looseness=-1

\subsection{Denoising Diffusion Models}

Denoising Diffusion Models (DDMs), rooted in nonequilibrium thermodynamics principles~\cite{sohl2015deep}, have emerged as a pioneering class of generative models, delivering outstanding outcomes across diverse domains including computer vision~\cite{luo2021diffusion, rombach2022high,ramesh2021zero,jin2024closed,li2025uniscene,li2025occscene}, natural language processing~\cite{li2022diffusion, gong2022diffuseq, reid2022diffuser}, and AI for science research~\cite{luo2021predicting, xu2022geodiff}. DDMs approximate the empirical data distribution through an iterative denoising procedure akin to score-based generative models~\cite{shao2022diffustereo}, which generate samples using Langevin dynamics informed by estimated gradients of the data distribution~\cite{jiang2022conditional}.

The standard diffusion models can be adapted for predictive tasks by incorporating specific guiding conditions. By controlling their behavior through such conditions, the performance of prediction tasks can be optimized. For instance, SR3~\cite{saharia2022image} generates photorealistic high-resolution images by conditioning on low-resolution inputs for super-resolution tasks. Similarly, DiffRF~\cite{muller2022diffrf} synthesizes radiance fields using posed images as additional conditions, resolving ambiguities with a rendering loss. Unlike DiffRF's one-to-many mapping, we employ the diffusion process as a one-to-one mapping to exploit geometry priors for precise and reliable 3D scene understanding.
DiffuStereo~\cite{shao2022diffustereo} utilizes an iterative diffusion model to derive highly accurate depth maps for automatic, high-fidelity human reconstruction from sparse-view inputs as conditions.
However, DiffuStereo directly refines depth maps using additional coarse information without fully exploiting geometric details. In contrast, we condition the diffusion model on explicit fine-grained geometric cues to reduce matching ambiguity and boundary over-smoothing during the diffusion process.

\begin{figure*}[!ht]
	\begin{center}
		\includegraphics[width=0.9\linewidth]{\assetpath 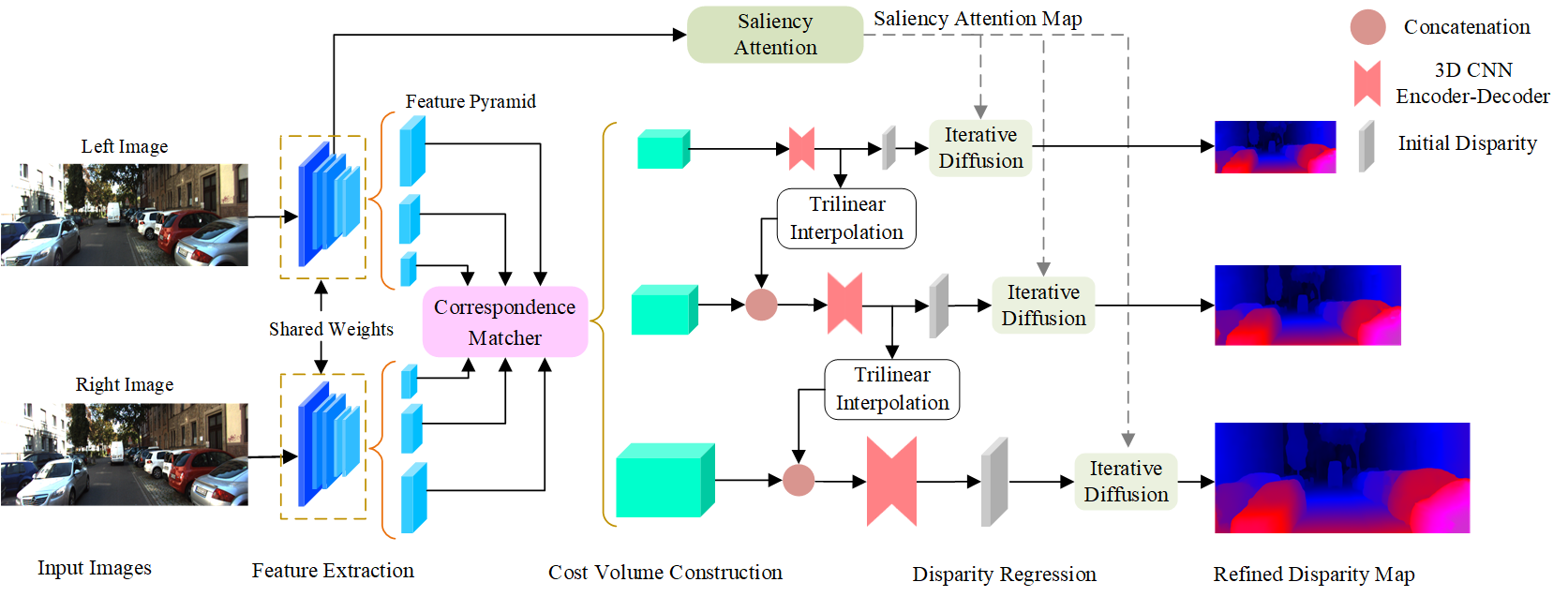}   
		\caption{Overall architecture of StereoDiffuer. The framework consists of four main components: feature-pyramid extraction, multi-scale cost-volume construction, initial disparity regression, and saliency-conditioned diffusion refinement. The cost-volume pyramid provides a multi-scale geometric prior for initial disparity estimation. In parallel, the Saliency Attention Perception (SAP) module extracts detail-aware saliency features from the input image features. A confidence map derived from the cost volume adaptively selects reliable saliency information, which is combined with the initial disparity map to form the collaborative condition. Finally, the conditional diffusion \re{U-Net} progressively refines the initial disparity map under the guidance of this condition.}
		\label{fig:1}                                 
	\end{center}                                 
\end{figure*}

\section{Method}

In this section, we present the proposed method for stereo matching. As illustrated in Figure~\ref{fig:1}, the overall framework consists of feature extraction, cost volume construction, disparity regression and diffusion-based iterative refinement. The details of the framework components are described below.   

\subsection{Feature Pyramid Extraction}
Given left ($I^{L}$) and right ($I^{R}$) input images with resolution $H\times W$, we first apply a shared feature extractor to construct feature-map pyramids. The feature map at the finest scale is generated by the backbone, while the other two scales are obtained using average-pooling operations with strides 2 and 4, respectively. These pyramids are then used to construct the cost-volume pyramid. The resolution at scale $s$ is $\frac{H}{2^{(s+1)}}\times\frac{W}{2^{(s+1)}}$, where $s\in\{1,2,3\}$.

\subsection{Cost Volume Pyramid Construction}

Multi-scale modeling has been extensively studied in low-level vision tasks, where hierarchical or multi-branch representations are commonly used to capture both global structures and local details. 
For image restoration, MB-TaylorFormer V2~\cite{jin2025mb} introduces a multi-branch linear Transformer with multi-scale patch embedding to model coarse-to-fine features with different receptive fields. 
For adverse-weather restoration, GridFormer~\cite{wang2024gridformer} adopts a grid-structured residual dense Transformer to enhance feature aggregation under diverse degradations. 
In desnowing, DDMSNet~\cite{zhang2021deep} exploits dense multi-scale representations together with semantic and geometric priors to recover clean images. 
For video deraining, ESTINet~\cite{zhang2022enhanced} further strengthens spatio-temporal interaction learning across consecutive frames. 
These studies show the general benefit of multi-scale modeling for recovering degraded visual details. 
Different from these image restoration methods, stereo matching requires explicit geometric reasoning along the disparity dimension. 
Therefore, we construct a multi-scale cost-volume pyramid in the disparity-aware matching space, where coarser volumes provide broader geometric context for ambiguous and textureless regions, while finer volumes preserve local disparity details around object boundaries and thin structures.

We construct a cost-volume pyramid to exploit multi-scale information, following~\cite{chang2018pyramid,cfnet}. Specifically, a ``correspondence matcher'' uses feature concatenation~\cite{chang2018pyramid} and group-wise correlation~\cite{guo2019group} between corresponding features at each scale. The cost-volume dimensions are $\frac{D}{4}\times\frac{H}{4}\times\frac{W}{4}$, $\frac{D}{8}\times\frac{H}{8}\times\frac{W}{8}$, and $\frac{D}{16}\times\frac{H}{16}\times\frac{W}{16}$, respectively. We then perform coarse-to-fine regularization to exchange context among the multi-scale cost volumes. The coarsest cost volume is first processed by a 3D CNN encoder--decoder along the spatial and disparity dimensions. Its regularized output is upsampled to the next pyramid scale, concatenated with the corresponding initial volume, and processed by another encoder--decoder regularization operation. This process continues until the finest cost volume is regularized, as follows:

\begin{equation} \label{eX:1}
    C_{i}^{A} = \Gamma\left \{ \mu_{T} \cdot  \left (C_{i-1}^{A}   \right )  \uplus C_{i}\right \},
\end{equation}
where $C_{i}^{{A}}$ and $C_{i-1}^{A}$ denote the regularized cost volume at the current and the previous scale, respectively. $C_{i}$ is the initial cost volume at the current scale. $\mu_{T}$ is a trilinear upsampling operation, and $\Gamma$ denotes the 3D CNN encoder-decoder regularization operation. $\uplus$ is the concatenation operation. It is worth noting that $C_{i\leq 0}^{A}=0$ and $C_{0}=0$.

\subsection{Disparity Regression}

At each scale of the regularized cost volume, we apply a softmax function to convert the cost volume into a probability volume $P(c)$. The $softmax$ operation is defined as:

\begin{equation} \label{eX:2}
{P(c_i)= \frac{\exp(c_i)}{\sum_{j=0}^{D_{max}-1}\exp(c_j)},}
\end{equation}
next, we apply soft-argmin at each scale of the probability volume to regress a disparity pyramid. The disparity regression operation is expressed as follows:
\begin{equation} \label{eX:3}
    {d^{pred} = \sum_{d=0}^{D_{max}-1} d \times P(c_{d}),}
\end{equation}
where $c_{d}$ is the regularized matching cost of the disparity $d$. 
To obtain disparity maps with the same resolution as the input image, we apply bilinear upsampling to the regressed disparity maps. 

Nevertheless, upsampling coupled with 3D-CNN encoder--decoder regularization can produce inaccurate disparities around thin objects, edges, and sharp object boundaries. To address this problem, we introduce the refinement strategy described in the next subsection to correct the resulting ``imperfect'' initial disparity map.

\begin{figure}[!ht]
	\begin{center}
		\includegraphics[width=0.99\linewidth]{\assetpath 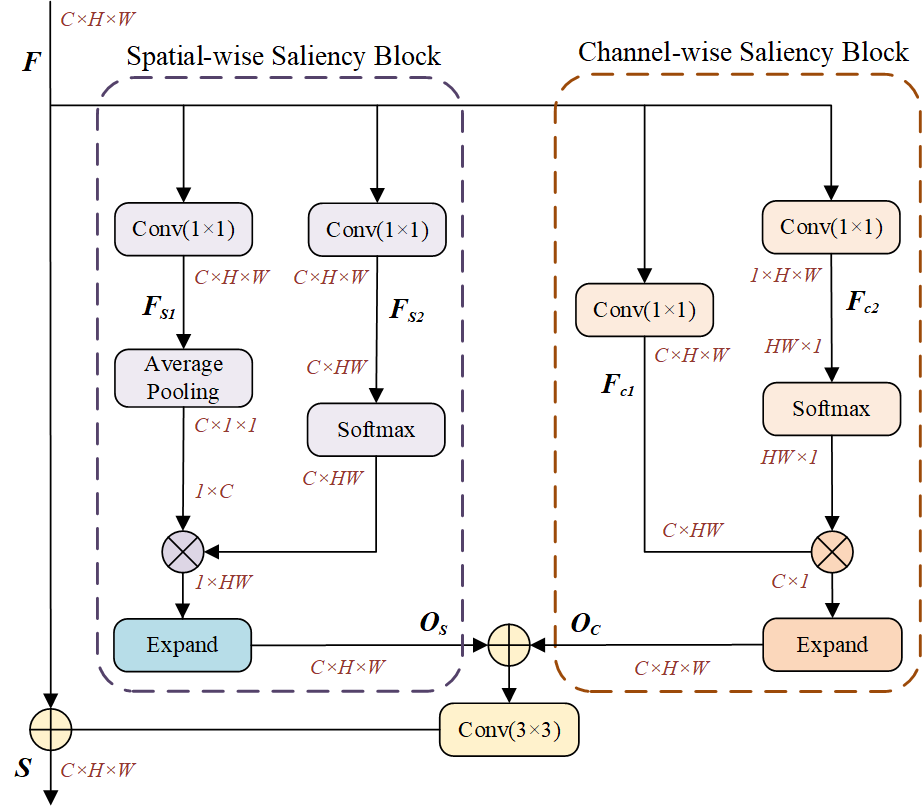}    
\caption{Architecture of the Saliency Attention Perception (SAP) module. 
SAP contains a spatial-wise saliency block (SSB) and a channel-wise saliency block (CSB). 
The SSB estimates pixel-level importance to emphasize spatially distinctive structures, while the CSB reweights feature channels to enhance geometry-related responses. 
The two saliency maps are fused and combined with the input feature to produce the final saliency attention map, which highlights boundaries, thin structures, and other high-frequency geometric details for disparity refinement.} 
		\label{fig:2}                                 
	\end{center}                                 
\end{figure}

\begin{figure}[!ht]
	\begin{center}
		\includegraphics[width=0.99\linewidth]{\assetpath 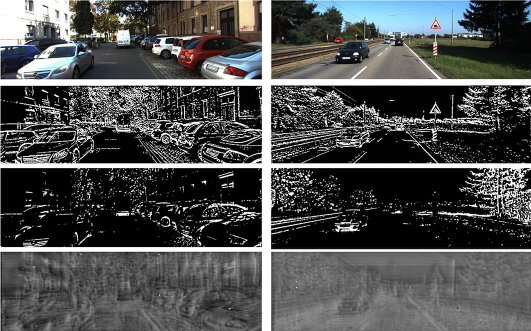}    
		\caption{Visualization of outputs from different attention modules. From top to bottom: input images, saliency attention maps from SAP, and attention maps from GCNet and NL-Net, respectively.\looseness=-1} 
		\label{fig:3}                                 
	\end{center}                                 
\end{figure}

\subsection{Refinement with Conditional Denoising Diffusion and Saliency Attention Perception}\label{section3.4}
To further improve disparity estimation quality in challenging regions, we propose to apply continuous diffusion-based refinement with the saliency attention perception.
Specifically, the refinement stage consists of a \textit{saliency attention perception module} and a \textit{conditional denoising diffusion module}. The former extracts saliency attention maps containing fine-grained geometric cues; these maps are combined with the initial disparity estimate to construct the condition for the diffusion module.

The refinement stage is designed to complement the preceding stereo matching blocks rather than replace them. 
The cost-volume pyramid and disparity regression provide an initial disparity map with coarse geometric consistency, but the regularization and upsampling operations may smooth out high-frequency details around boundaries and thin structures. 
SAP therefore extracts saliency-aware geometric cues from the image features, while the cost-volume confidence indicates the reliability of the initial matching result. 
By combining the initial disparity map and confidence-guided saliency features as the condition of the diffusion model, the reverse denoising process progressively corrects residual disparity errors in challenging geometric regions.

\subsubsection{Saliency Attention Perception}
To guide refinement of the initial disparity maps, we introduce the Saliency Attention Perception (SAP) module to extract detailed geometric knowledge and fine-grained context. Unlike attention mechanisms that primarily aggregate global spatial correlations~\cite{wang2018non,cao2019gcnet}, SAP learns pixel- and channel-level importance to emphasize geometric details such as edges and sharp boundaries while suppressing less informative low-frequency responses.
In particular, we feed the feature map from the ResNet layer into the saliency attention perception (SAP) module, which consists of two blocks: spatial-wise saliency block (SSB) and channel-wise saliency block (CSB). 

\noindent\textbf{Spatial-wise Saliency.} The spatial-wise saliency block applies two parallel $1\times1$ convolutions to the input feature map $\textbf{F}\in\mathbb{R}^{C\times H\times W}$ to generate $\textbf{F}_{s1},\textbf{F}_{s2}\in\mathbb{R}^{C\times H\times W}$. An average-pooling operation, which serves as a channel descriptor, is applied to $\textbf{F}_{s1}$. We reshape $\textbf{F}_{s2}$ to $\mathbb{R}^{C\times HW}$ and apply softmax to encode pixel-level importance within each channel. The pooled output is reshaped to $\mathbb{R}^{1\times C}$ and multiplied by the softmax output to generate the spatial saliency map. SSB is expressed as follows:
\begin{equation} \label{eX:4}
	\textbf{O}_{s}=\xi_{e}  \left\{  \mathbb{R}_{1} \left( \mathbb{P}_{avg} \left(\textbf{F}_{s1 }  \right) \right)  \otimes   \mathbb{S} \left( \mathbb{R}_{2} \left(\textbf{F}_{s2 } \right) \right) \right\},
\end{equation}
where $\xi_{e}$ is an expansion operation. $\mathbb{R}_{1}$ and $\mathbb{R}_{2}$ denote reshape operations. $\mathbb{P}_{avg}$ denotes average pooling. 
$\otimes$ represents matrix multiplication and $\mathbb{S}$ is a softmax operation\looseness=-1.

\noindent\textbf{Channel-wise Saliency.} Similar to SSB, CSB applies two $1\times1$ convolutions to generate $\textbf{F}_{c1}\in\mathbb{R}^{C\times H\times W}$ and $\textbf{F}_{c2}\in\mathbb{R}^{1\times H\times W}$. We reshape $\textbf{F}_{c2}$ to $\mathbb{R}^{HW\times1}$ and apply softmax to obtain spatial weights. Before multiplication, $\textbf{F}_{c1}$ is reshaped to $\mathbb{R}^{C\times HW}$. Their product gives the channel-wise saliency map $\textbf{O}_{c}$:
\begin{equation} \label{eX:5}
    \textbf{O}_{c}=\xi_{e}  \left\{  \mathbb{R}_{1} \left( \textbf{F}_{c1}   \right)  \otimes  \mathbb{S} \left( \mathbb{R}_{2} (\textbf{F}_{c2}  ) \right)  \right\}.
\end{equation}

\begin{figure*}[!ht]
\vspace{-0pt}
\centering
\rcap
\includegraphics[width=0.9\linewidth]{\assetpath 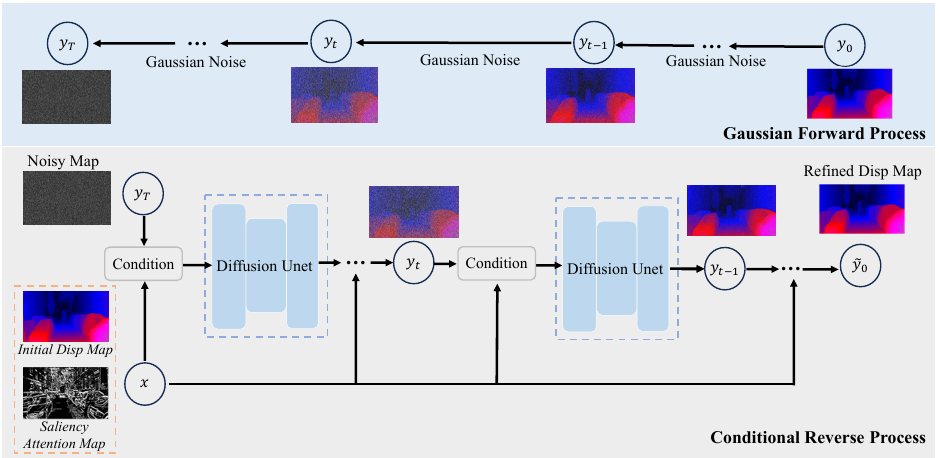}
\vspace{-0pt}
\caption{Architectural details of the conditional denoising diffusion module. In the Gaussian forward process, the target disparity representation $\boldsymbol{y}_{0}$ is gradually corrupted over $T$ time steps. In the conditional reverse process, the diffusion U-Net progressively predicts the clean target $\hat{\boldsymbol{y}}_{0}$. The initial disparity map and confidence-guided SAP feature are concatenated to construct the collaborative condition $\boldsymbol{x}$ used throughout the reverse process.}
\label{overall}
\vspace{-0pt}
\end{figure*}

Next, we apply element-wise summation to fuse the spatial-wise and channel-wise saliency maps. Finally, we employ a 2D convolution to process the fusion output and sum it with the input feature map $\textbf{F}$ to compute the saliency attention map $\textbf{S}$:
\begin{equation} \label{eX:6}
    \textbf{S}= W  ( \textbf{O}_{s} + \textbf{O}_{c} ) + \textbf{F},
\end{equation}
where $W$ denotes the convolution layer with the kernel size of $3\times3$.
We visualize the saliency attention maps obtained from our approach and the outputs of representative attention-based methods (GCNet~\cite{cao2019gcnet} and NL-Net~\cite{wang2018non}) in Figure~\ref{fig:3}. \re{As shown, SAP highlights finer geometric structures in the displayed examples.}

\subsubsection{Conditional Denoising Diffusion}

To progressively refine the initial ``imperfect'' disparity maps, we use an iterative denoising diffusion model to decompose the correction into a sequence of geometry updates.
Standard generative diffusion models learn one-to-many mappings through forward and reverse processes~\cite{saharia2022image}.
In our setting, a diffusion \re{U-Net}~\cite{saharia2022image} learns a conditional one-to-one mapping to the target disparity representation.
In the forward process, the clean target $\boldsymbol{y}_{0}$ is progressively corrupted over $T$ time steps.
In the reverse process, the \re{U-Net} estimates the clean target from a noisy input under the guidance of the collaborative condition $\boldsymbol{x}$.

\noindent\textbf{Gaussian Forward Process.} \label{sec311}
Given a ground-truth disparity representation $\boldsymbol{y}_{0}$, the forward process follows a discrete-time Markov chain:
\begin{equation}
q\left(\boldsymbol{y}_{1:T} \mid \boldsymbol{y}_0\right)=\prod_{t=1}^T q\left(\boldsymbol{y}_t \mid \boldsymbol{y}_{t-1}\right),
\end{equation}
\begin{equation}
q\left(\boldsymbol{y}_t \mid \boldsymbol{y}_{t-1}\right)=\mathcal{N}\left(\boldsymbol{y}_t \mid \sqrt{\alpha_t}\boldsymbol{y}_{t-1},\left(1-\alpha_t\right)\boldsymbol{I}\right),
\end{equation}
where $\alpha_t=1-\beta_t$, $\beta_t$ is the noise variance at step $t$, $\mathcal{N}$ denotes a multivariate normal distribution, and $\boldsymbol{I}$ is the \re{identity matrix}.
Marginalizing the intermediate steps gives
\begin{equation}
q\left(\boldsymbol{y}_t \mid \boldsymbol{y}_0\right)=\mathcal{N}\left(\boldsymbol{y}_t\mid\sqrt{\bar{\alpha}_t} \boldsymbol{y}_0,\left(1-\bar{\alpha}_t\right) \boldsymbol{I}\right) ,
\end{equation}
where $\bar{\alpha}_t=\prod_{i=1}^t\alpha_i$.
\re{In all experiments, we set $T=1000$ and use a cosine noise schedule. To distinguish the zero-based implementation index from the one-based Markov-chain notation above, let $j\in\{0,\ldots,T-1\}$ index the stored schedule. We define
\begin{equation}\label{eq:cosine_base}
f(u)=\cos^{2}\!\left(\frac{\pi(u+s)}{2(1+s)}\right),
\qquad s=0.008.
\end{equation}
The variance coefficient at index $j$ is
\begin{equation}\label{eq:cosine_schedule}
\beta_j=\operatorname{clip}\!\left(
1-\frac{f((j+1)/T)}{f(j/T)},10^{-4},0.9999
\right),
\end{equation}
with $\alpha_j=1-\beta_j$ and $\widetilde{\alpha}_j=\prod_{i=0}^{j}\alpha_i$. During training, a schedule index is sampled independently for each example as $j\sim\mathcal{U}\{0,\ldots,T-1\}$, followed by $\boldsymbol{\epsilon}\sim\mathcal{N}(\boldsymbol{0},\boldsymbol{I})$. The corresponding noisy state is generated in closed form as
\begin{equation}\label{eq:forward_sample}
\boldsymbol{y}^{[j]}=\sqrt{\widetilde{\alpha}_{j}}\boldsymbol{y}_{0}
+\sqrt{1-\widetilde{\alpha}_{j}}\boldsymbol{\epsilon}.
\end{equation}}
The posterior distribution of ${y}_{t-1}$ can be further derived by exploiting algebraic manipulation 
with $({y}_{0},{y}_{t})$:
\begin{equation}
 q\left(\boldsymbol{y}_{t-1} \mid \boldsymbol{y}_0, \boldsymbol{y}_t\right)=\mathcal{N}\left(\boldsymbol{y}_{t-1} \mid \boldsymbol{\mu}, \sigma^2 \boldsymbol{I}\right) 
\end{equation}
where $\boldsymbol{\mu}=\frac{\sqrt{\bar{\alpha}_{t-1}}\left(1-\alpha_t\right)}{1-\bar{\alpha}_t} \boldsymbol{y}_0+\frac{\sqrt{\alpha_t}\left(1-\bar{\alpha}_{t-1}\right)}{1-\bar{\alpha}_t} \boldsymbol{y}_t $, and 
$\sigma^2=\frac{\left(1-\bar{\alpha}_{t-1}\right)\left(1-\alpha_t\right)}{1-\bar{\alpha}_t}$.

\noindent\textbf{Conditional Reverse Process.}
The conditional reverse process iteratively denoises the input to recover $\boldsymbol{y}_{0}$.
Each reverse transition is defined as~\cite{saharia2022image}
 \begin{equation}
 p_\theta\left(\boldsymbol{y}_{0: T} \mid \boldsymbol{x}\right)=p\left(\boldsymbol{y}_T\right) \prod_{t=1}^{T} p_\theta\left(\boldsymbol{y}_{t-1} \mid \boldsymbol{y}_t, \boldsymbol{x}\right),
\end{equation}

\begin{equation}
p_\theta\left(\boldsymbol{y}_{t-1} \mid \boldsymbol{y}_t, \boldsymbol{x}\right)=\mathcal{N}\left(\boldsymbol{y}_{t-1} \mid \mu_\theta\left( \boldsymbol{y}_t, \bar{\alpha}_t, \boldsymbol{x} \right), \sigma_t^2 \boldsymbol{I}\right) ,
\end{equation}
\re{Here, $\boldsymbol{x}$ denotes the collaborative condition, including the initial disparity map and confidence-guided saliency feature. In the zero-based implementation, the noisy state $\boldsymbol{y}^{[j]}$ and $\boldsymbol{x}$ are concatenated along the channel dimension and fed into the diffusion U-Net. We use direct clean-target ($\boldsymbol{y}_0$) prediction rather than noise or velocity prediction:
\begin{equation}\label{eq:x0_prediction}
\hat{\boldsymbol{y}}_{0}=f_{\theta}(\boldsymbol{y}^{[j]},j,\boldsymbol{x}).
\end{equation}
The clean-target estimate is supervised against the ground-truth disparity representation, while the collaborative condition is held fixed for all sampled schedule indices.}

\re{At inference, we use deterministic sparse reverse sampling with $\eta=0$. For the default four-step setting, the selected zero-based schedule indices are
\begin{equation}\label{eq:four_step_indices}
\left\lfloor\operatorname{linspace}\!\left(0,\sqrt{0.8T},4\right)^{2}\right\rfloor
=\{0,88,355,800\},
\end{equation}
which are traversed as $800\rightarrow355\rightarrow88\rightarrow0$. Starting from Gaussian noise $\boldsymbol{y}^{[800]}$, the model predicts $\hat{\boldsymbol{y}}_{0}$ at every selected index. For each successive pair $j>j'$, the implied noise is recovered by
\begin{equation}\label{eq:implied_noise}
\hat{\boldsymbol{\epsilon}}_{j}=
\frac{\boldsymbol{y}^{[j]}-\sqrt{\widetilde{\alpha}_{j}}\hat{\boldsymbol{y}}_{0}}
{\sqrt{1-\widetilde{\alpha}_{j}}}.
\end{equation}
The deterministic update to $j'$ is
\begin{equation}\label{eq:sparse_reverse_update}
\boldsymbol{y}^{[j']}=\sqrt{\widetilde{\alpha}_{j'}}\hat{\boldsymbol{y}}_{0}
+\sqrt{1-\widetilde{\alpha}_{j'}}\hat{\boldsymbol{\epsilon}}_{j}.
\end{equation}
The collaborative condition is reused at every selected index. The model evaluation at $j=0$ produces the final clean-target prediction, which is converted into the full-resolution disparity map. Thus, the default strategy requires four U-Net evaluations in total.}


\noindent\textbf{{Collaborative Condition Construction.}}
To combine the complementary information in the saliency attention map and the initial disparity map, we construct a collaborative condition $\boldsymbol{x}$ to guide iterative denoising diffusion.

In particular, we first establish a confidence map $\textbf{C}$ by identifying the highest probability across all disparity-hypothesis planes of the cost volume $\textbf{V}$.
Next, we adaptively extract saliency attention information according to the confidence values in $\textbf{C}$ by applying linear cross-attention~\cite{kitaev2020reformer,shen2021efficient}, which is formed as:
\begin{equation} \label{eqca}
\begin{split}
\textbf{C}' &= CrossAtt(Q,K,V )  \\
                    & = \phi_q(Q) (\phi_k{ (K) }^{T} V ),
\end{split}
\end{equation}
where $\textbf{C}'$ represents the extracted saliency feature map. 
The query $Q$ is derived from the confidence map $\textbf{C}$, while the key $K$ and value $V$ are generated from the saliency attention map $\textbf{S}$.
$\phi_q$ and $\phi_k$ denote softmax operations along the rows and columns of the input matrix, respectively. In this way, reliable saliency information is selected according to matching-cost confidence and used to supplement the initial disparity estimate.

Subsequently, the extracted saliency feature map $\textbf{C}'$ is concatenated with the initial disparity map $\textbf{D}_{ini}$ along the channel dimension to generate the collaborative condition $\boldsymbol{x}$ as:
\begin{equation} \label{eq14}
\boldsymbol{x} = \operatorname{Cat}\left(\textbf{C}',\textbf{D}_{ini};\,\mathrm{dim}=C\right),
\end{equation}

Through collaborative condition construction, confidence-guided saliency features supplement the initial disparity map with geometric details throughout iterative denoising diffusion.

\subsection{Loss Function}
The entire network is trained in an end-to-end manner with an MSE loss. Specifically, we adopt the strategy in~\cite{chang2018pyramid,cfnet} to compute the total loss as a weighted summation of the three disparity map outputs during the training phase: 
\begin{equation} \label{eX:15}
   L_{all} = \sum_{k=0}^{2}\eta _{k} \cdot \left\{ \operatorname{MSE}(\textbf{D}_{ini}^{k},D^{gt}) + \operatorname{MSE}(\textbf{D}_{ref}^{k},D^{gt}) \right\},
\end{equation}
where $\eta _{k}$ is the $k$-th loss coefficient, which is assigned as 0.5, 0.7, and 1.0, respectively. $\textbf{D}_{ini}^{k}$ and $\textbf{D}_{ref}^{k}$ represent the $k$-th initial and refined disparity maps, respectively. $D^{gt}$ denotes the ground-truth disparities.
Note that in the test phase, we use the predictions at the finest scale as the final disparity maps.

\section{Experiments}

In this section, we present the implementation details and the evaluations of the proposed network with a comprehensive set of experiments on standard stereo matching benchmarks.

\subsection{Implementation Details}
\noindent \textbf{Datasets.} Scene Flow~\cite{mayer2016large} is a large synthetic dataset containing 35,454 training stereo pairs and 4,370 test stereo pairs with ground-truth disparities. KITTI 2012~\cite{geiger2012we} and KITTI 2015~\cite{menze2015object} are real-world driving datasets. KITTI 2012 contains 194 training pairs and 195 test pairs; we use 160 of the provided training pairs for training and the remaining 34 for validation. KITTI 2015 contains 200 training pairs and 200 test pairs; we use 160 of the provided training pairs for training and the remaining 40 for validation.

\noindent \textbf{Training.} We implement the network in \re{PyTorch} and train it on an NVIDIA RTX 4090 GPU with a batch size of 6. We use Adam ($\beta_{1}=0.9$, $\beta_{2}=0.999$) as the optimizer. Network weights are randomly initialized, and image intensities are normalized to $[-1,1]$. During training, input image pairs are randomly cropped to $256\times512$. The maximum disparity is 192, and pixels above this threshold are excluded from the loss. We first train on Scene Flow for 20 epochs with an initial learning rate of 0.001, halved at epochs 14, 16, and 18. We then jointly fine-tune on the mixed KITTI 2012 and KITTI 2015 training splits for 300 epochs, using a learning rate of 0.001 for the first 200 epochs and 0.0001 thereafter. Finally, we fine-tune separately on KITTI 2012 and KITTI 2015 for 500 epochs, reducing the learning rate from 0.001 to 0.0001 after 200 epochs.

\begin{table*}[!ht]
\begin{center}
\caption{
Quantitative results on the KITTI 2012 and KITTI 2015 test sets. 
For KITTI 2012, the evaluation metrics report the percentages of erroneous pixels at thresholds of 2 and 3 pixels, as well as the average disparity error in non-occluded regions (Noc) and all regions (All). 
For KITTI 2015, the metrics report the percentage of disparity outliers in the first frame (D1) over background regions (bg), foreground regions (fg), and all ground-truth pixels (all).
}
\renewcommand\tabcolsep{15.0pt}
\resizebox{0.99\textwidth}{!}{
\begin{tabular}{l|cccccc|ccc}
\toprule
 & \multicolumn{6}{c|}{KITTI 2012} & \multicolumn{3}{c}{KITTI 2015} \\ 
\cline{2-10}   
\multirow{2}{*}{Method}  &
  \multicolumn{2}{c}{\re{2-pixel error (\%)}} &
  \multicolumn{2}{c}{\re{3-pixel error (\%)}} &
  \multicolumn{2}{c}{Mean Error} &
  \multicolumn{3}{c}{\re{D1 (\%)}} 
  \\  
\cline{2-10} 
& Noc & All & Noc & All & Avg-Noc & Avg-All
& D1-bg & D1-fg & D1-all \\
\midrule
GCNet~\cite{kendall2017end} 
& 2.71 & 3.46 & 1.77 & 2.30 & 0.6 & 0.7 & 2.21 & 6.16 & 2.87 \\ 

PSMNet~\cite{chang2018pyramid}  
& 2.44 & 3.01 & 1.49 & 1.89 & 0.5 & 0.6 & 1.86 & 4.62 & 2.32 \\ 

GwcNet~\cite{guo2019group} 
& 2.16 & 2.71 & 1.32 & 1.70 & 0.5 & \textbf{0.5} & 1.74 & 3.93 & 2.11 \\ 

GANet-deep~\cite{zhang2019GANet} 
& 1.89 & 2.50 & 1.19 & 1.60 & \textbf{0.4} & \textbf{0.5} & 1.48 & 3.46 & 1.81 \\ 

StereoDRNet~\cite{chabra2019stereodrnet} 
& 2.29 & 2.87 & 1.42 & 1.83 & 0.5 & \textbf{0.5} & 1.72 & 4.95 & 2.26 \\ 

AANet~\cite{xu2020aanet} 
& 2.90 & 3.60 & 1.91 & 2.42 & 0.5 & 0.6 & 1.99 & 5.39 & 2.55 \\ 

LEAStereo~\cite{cheng2020hierarchical} 
& 1.90 & 2.39 & 1.13 & \textbf{1.45} & 0.5 & \textbf{0.5} & 1.40 & 2.91 & 1.65 \\  

SGNet~\cite{chen2020sgnet} 
& 2.22 & 2.89 & 1.38 & 1.85 & 0.5 & \textbf{0.5} & 1.63 & 3.76 & 1.99 \\ 

EdgeStereo~\cite{song2020edgestereo}  
& 2.32 & 2.88 & 1.46 & 1.83 & \textbf{0.4} & \textbf{0.5} & 1.84 & 3.30 & 2.08 \\ 

HITNet~\cite{tankovich2021hitnet} 
& 2.00 & 2.65 & 1.41 & 1.89 & \textbf{0.4} & \textbf{0.5} & 1.74 & 3.20 & 1.98 \\ 

BGNet+~\cite{bgnet2021}   
& 2.78 & 3.35 & 1.62 & 2.03 & 0.5 & 0.6 & 1.81 & 4.09 & 2.19 \\ 

RAFT-Stereo~\cite{lipson2021raft} 
& -- & -- & -- & -- & -- & -- 
& 1.75 & 2.89 & 1.96 \\ 

ACVNet~\cite{xu2022attention} 
& 1.83 & 2.34 & 1.13 & 1.47 & \textbf{0.4} & \textbf{0.5} 
& \textbf{1.37} & 3.07 & 1.65 \\ 

CVCNet~\cite{guo2022cvcnet} 
& -- & -- & 1.21 & 1.57 & 0.5 & \textbf{0.5} 
& 1.45 & 3.22 & 1.74 \\ 

DKT-IGEV~\cite{zhang2024robust}  
& -- & -- & 1.22 & 1.56 & -- & -- & 1.46 & 3.05 & 1.72 \\ 

Selective-RAFT~\cite{wang2024selective} 
& -- & -- & -- & -- & -- & -- 
& 1.41 & 2.71 & 1.63 \\ 

\rowcolor{Gray} StereoDiffuer (ours) 
& \textbf{1.78} & \textbf{2.31} & \textbf{1.12} & 1.49 
& \textbf{0.4} & \textbf{0.5} 
& 1.39 & \textbf{2.50} & \textbf{1.57} \\ 
\bottomrule
\end{tabular}
}
\label{tab:4}
\end{center}
\end{table*}

\begin{figure*}[!t]
	\begin{center}
		\includegraphics[width=0.9\linewidth]{\assetpath 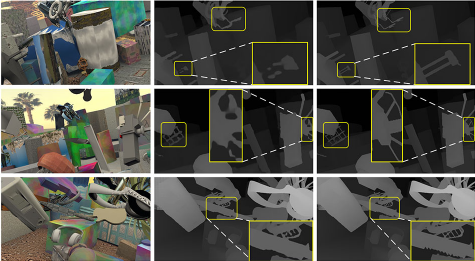}    
		\caption{Visualization of refinement results on the Scene Flow test set. From left to right: input images, initial disparity maps, and refined disparity maps. In these examples, the refinement produces sharper object boundaries.}  
		\label{fig:5}   
	\end{center}     
\end{figure*}

\begin{figure*}[!ht]
	\begin{center}
		\includegraphics[width=0.9\linewidth]{\assetpath 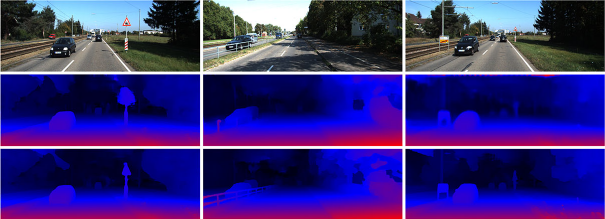}    
		\caption{Visualization of refinement results on the KITTI 2015 validation set. From top to bottom: input images, initial disparity maps, and refined disparity maps. In these examples, the refinement preserves thin structures and produces sharper object boundaries.}  
		\label{fig:4}                                 
	\end{center}                                 
\end{figure*}

\subsection{\re{Performance}}


\noindent\textbf{Scene Flow.} 
We compare our network with representative stereo matching methods on the Scene Flow \re{test} set in Table~\ref{tab:3}, including GANet~\cite{zhang2019GANet}, AANet~\cite{xu2020aanet}, FADNet~\cite{wang2020fadnet}, LEAStereo~\cite{cheng2020hierarchical}, GwcNet~\cite{guo2019group}, BGNet+~\cite{bgnet2021}, CVCNet~\cite{guo2022cvcnet}, ACVNet~\cite{xu2022attention}, RAFT-Stereo~\cite{lipson2021raft}, IGEV-Stereo~\cite{xu2023iterative}, and Selective-RAFT~\cite{wang2024selective}. 
We use the end-point error (EPE) as the evaluation metric, which measures the average disparity error in pixels. 
\re{With 12 reverse steps for the final benchmark evaluation, StereoDiffuer obtains an EPE of 0.47, matching the reported EPEs of IGEV-Stereo and Selective-RAFT and yielding a lower EPE than the other methods listed in Table~\ref{tab:3}. Controlled ablations use the default four-step setting unless otherwise stated.} These results indicate that saliency-conditioned diffusion refinement provides effective disparity correction while maintaining competitive overall accuracy.\looseness=-1

\begin{figure}[!t]
	\begin{center}
		\rcap
		\includegraphics[width=0.99\linewidth]{\assetpath 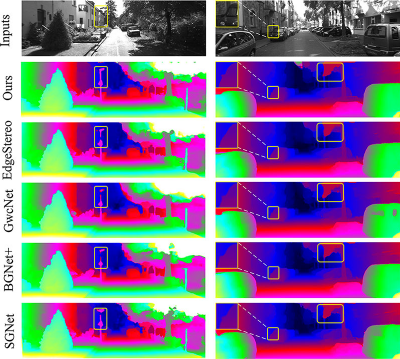}   
		\caption{Qualitative comparison with EdgeStereo, GwcNet, BGNet+, and SGNet on the KITTI 2012 test set. In these examples, StereoDiffuer preserves finer thin structures and object edges than the displayed comparison results.}
		\label{fig:6}                        
	\end{center}                             
\end{figure}

\begin{figure}[t]
	\begin{center}
		\rcap
		\includegraphics[width=0.99\linewidth]{\assetpath 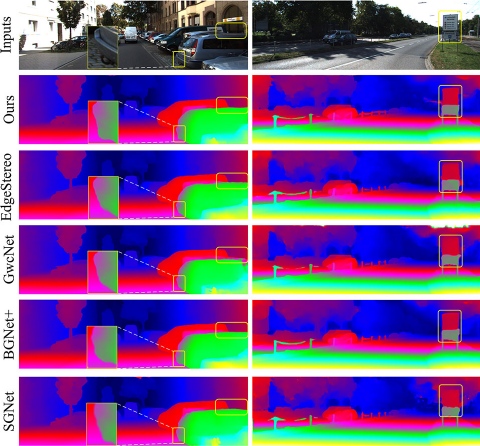}   
		\caption{Qualitative comparison with EdgeStereo, GwcNet, BGNet+, and SGNet on the KITTI 2015 test set. In these examples, StereoDiffuer produces sharper object boundaries and more coherent disparity estimates in low-texture regions.} 
		\label{fig:7}                        
	\end{center}                              
\end{figure}


\noindent\textbf{KITTI.} 
We conduct extensive experiments on the KITTI 2012 and KITTI 2015 datasets. 
The quantitative comparisons are presented in Table~\ref{tab:4}, with recent baselines, including RAFT-Stereo~\cite{lipson2021raft}, ACVNet~\cite{xu2022attention}, CVCNet~\cite{guo2022cvcnet}, DKT-IGEV~\cite{zhang2024robust}, and Selective-RAFT~\cite{wang2024selective}. 
\re{On KITTI 2012, StereoDiffuer reports the lowest 2-pixel errors and the lowest non-occluded 3-pixel error among the methods listed in Table~\ref{tab:4}. On KITTI 2015, it obtains a D1-all error of 1.57\% and the lowest listed foreground error, with D1-fg at 2.50\%. These measurements are consistent with the design goal of saliency-conditioned refinement in foreground and boundary regions.}

Figures~\ref{fig:6} and~\ref{fig:7} provide qualitative comparisons with \re{EdgeStereo}~\cite{song2020edgestereo}, GwcNet~\cite{guo2019group}, BGNet+~\cite{bgnet2021}, and SGNet~\cite{chen2020sgnet}. \re{In the displayed examples, StereoDiffuer preserves finer geometric details around thin structures and object boundaries.}

\noindent \textbf{Visualization of the refinement process.} We visualize refinement results on Scene Flow and KITTI in Figures~\ref{fig:5} and~\ref{fig:4}. The displayed results show corrections around edges, thin structures, and sharp object boundaries, consistent with the explicit modeling of geometric details and iterative diffusion refinement.

            


\begin{table}[!ht]
\centering
\caption{
Quantitative results on the Scene Flow \re{test set}. We use end-point error (EPE) as the evaluation metric. \re{The StereoDiffuer result uses 12 reverse steps; controlled ablations use four steps unless otherwise stated.}
}
\label{tab:3}
\renewcommand{\arraystretch}{1.05}
\renewcommand\tabcolsep{10pt}
\resizebox{0.99\linewidth}{!}{
\begin{tabular}{l c | l c}
\toprule
Method & EPE $\downarrow$ & Method & EPE $\downarrow$ \\
\midrule
GANet~\cite{zhang2019GANet} & 0.83 
& CVCNet~\cite{guo2022cvcnet} & 0.68 \\

AANet~\cite{xu2020aanet} & 0.87 
& ACVNet~\cite{xu2022attention} & 0.48 \\

FADNet~\cite{wang2020fadnet} & 0.84 
& RAFT-Stereo~\cite{lipson2021raft} & 0.56 \\

LEAStereo~\cite{cheng2020hierarchical} & 0.78 
& IGEV-Stereo~\cite{xu2023iterative} & \textbf{0.47} \\

GwcNet~\cite{guo2019group} & 0.76 
& Selective-RAFT~\cite{wang2024selective} & \textbf{0.47} \\

BGNet+~\cite{bgnet2021} & 0.63 
& \cellcolor{Gray}StereoDiffuer (ours) & \cellcolor{Gray}\textbf{0.47} \\
\bottomrule
\end{tabular}
}
\end{table}

\noindent\textbf{Performance in challenging edge regions.}
\re{We conduct the edge-region evaluation on the 40-pair KITTI 2015 validation split used in our ablation studies. Let $\mathcal{V}=\{\boldsymbol{p}\mid 0<d^{gt}(\boldsymbol{p})<192\}$ denote the valid-pixel set. For each pixel with valid horizontal and vertical neighbors, we compute
\begin{equation}\label{eq:edge_gradient}
g(\boldsymbol{p})=\max\!\left(
\begin{aligned}
&\left|d^{gt}(\boldsymbol{p}+\boldsymbol{e}_{x})-d^{gt}(\boldsymbol{p})\right|,\\[-1mm]
&\left|d^{gt}(\boldsymbol{p}+\boldsymbol{e}_{y})-d^{gt}(\boldsymbol{p})\right|
\end{aligned}
\right).
\end{equation}
The initial disparity-discontinuity mask is defined by $g(\boldsymbol{p})>1$ pixel. We dilate this mask once with a $3\times3$ square structuring element and intersect the result with $\mathcal{V}$ to obtain the final edge region. The non-edge region is the complement of the edge region within $\mathcal{V}$. All methods are evaluated using the same masks and full-resolution predictions, without boundary-specific post-processing. EPE is the mean absolute disparity error over all pixels in the corresponding region, and the 3-pixel error is the percentage of pixels with an absolute disparity error greater than 3 pixels. Pixels are pooled over the complete validation split before each metric is calculated. As shown in Table~\ref{tab:edge_region}, StereoDiffuer obtains the lowest errors among the compared methods in the edge region, with an Edge EPE of \textbf{0.76} and an Edge 3-pixel error of \textbf{2.58\%}. The larger relative improvement in the edge region is consistent with the goal of saliency-conditioned boundary refinement.
}

\begin{table}[htbp]
\rcap
\centering
\caption{
Performance on the 40-pair KITTI 2015 validation split in edge and non-edge regions. Valid pixels satisfy $0<d^{gt}<192$. The initial edge mask is obtained by thresholding horizontal and vertical forward differences of valid ground-truth disparities at 1 pixel and is dilated once with a $3\times3$ square structuring element. All methods use the same masks and full-resolution predictions without boundary-specific post-processing. EPE and 3-pixel error are pooled over all valid pixels in each region.
}
\label{tab:edge_region}
\renewcommand\tabcolsep{7pt}
\resizebox{0.99\linewidth}{!}{
\begin{tabular}{lcccc}
\toprule
Method 
& \makecell{Edge\\EPE $\downarrow$}
& \makecell{Edge\\\re{3-pixel error (\%)} $\downarrow$}
& \makecell{Non-edge\\EPE $\downarrow$}
& \makecell{Non-edge\\\re{3-pixel error (\%)} $\downarrow$} \\
\midrule
GwcNet~\cite{guo2019group} 
& {1.52} & {5.12} & {0.68} & {1.75} \\
BGNet+~\cite{bgnet2021} 
& {1.34} & {4.65} & {0.58} & {1.62} \\
ACVNet~\cite{xu2022attention} 
& {1.15} & {3.84} & {0.45} & {1.35} \\
IGEV-Stereo~\cite{xu2023iterative} 
& {0.98} & {3.32} & {0.39} & {1.20} \\
Selective-RAFT~\cite{wang2024selective} 
& {0.94} & {3.21} & {0.37} & {1.15} \\
\midrule
\rowcolor{gray!20}
\re{StereoDiffuer (ours)} 
& {\textbf{0.76}} & {\textbf{2.58}} & {\textbf{0.35}} & {\textbf{1.12}} \\
\bottomrule
\end{tabular}
}
\end{table}

\begin{table*}[!ht]
\begin{center}
\caption{Ablation results for the proposed framework. The 3-pixel error is computed on the KITTI 2015 validation split, and end-point error (EPE) is computed on the Scene Flow test set. \re{These controlled experiments use four reverse diffusion steps unless otherwise stated.}}
\renewcommand\tabcolsep{12.0pt}
\resizebox{0.99\textwidth}{!}{
\begin{tabular}{ccc|cc|c|c|c|c}
\toprule
\multicolumn{3}{c|}{Cost Volume Pyramid}  & \multicolumn{2}{c|}{Saliency Attention Perception} 
&\multicolumn{1}{c|}{\multirow{2}{*}{\begin{tabular}[c]{@{}c@{}} Iterative\\ Diffusion\end{tabular}}} 
&\multicolumn{1}{c|}{\multirow{2}{*}{\begin{tabular}[c]{@{}c@{}}Iterative\\ GRU\end{tabular}}} 
&\multicolumn{1}{c|}{\multirow{2}{*}{\begin{tabular}[c]{@{}c@{}} KITTI 2015 \\ \re{3-pixel error (\%)}  \end{tabular}}} 
&\multicolumn{1}{c}{\multirow{2}{*}{\begin{tabular}[c]{@{}c@{}}  Scene Flow \\ EPE  \end{tabular}}} 
			
  \\  \cline{1-5}
  Large & Middle & Small & Spatial-wise Saliency & Channel-wise Saliency& \multicolumn{1}{c|}{}& \multicolumn{1}{c|}{}& \multicolumn{1}{c|}{} \\ \midrule
$\surd$ &  &   & $\surd$ & $\surd$ & $\surd$ & & 2.03 & 1.30 \\
$\surd$ &$\surd$ & &$\surd$ & $\surd$&$\surd$  & &{1.91}&{1.22}\\
$\surd$ &$\surd$  &$\surd$  & $\surd$ & $\surd$ & $\surd$  & & 1.81 & 1.08 \\ \midrule

$\surd$ & $\surd$ & $\surd$ & & & $\surd$ & & 1.74 & 0.90 \\ 
$\surd$ & $\surd$ & $\surd$ & \re{$\surd$} & & $\surd$ &  & {1.69} & {0.81} \\ 
$\surd$ & $\surd$ & $\surd$ & & $\surd$ & $\surd$  & & 1.63 & 0.76     \\
\midrule
$\surd$ & $\surd$ & $\surd$ & $\surd$ & $\surd$ & &$\surd$ 
 &{1.73} &{0.81}  \\ \midrule
$\surd$ & $\surd$ & $\surd$ & $\surd$ & $\surd$ & $\surd$  &&\textbf{1.52}&\textbf{0.51}  \\ \bottomrule
\end{tabular}
}
\label{tab:2} 
\end{center}
\end{table*}

\subsection{Ablation Study}
  
As shown in Table~\ref{tab:2}, we conduct ablation studies on the KITTI 2015 validation split and Scene Flow test set to evaluate the contribution of each component.\looseness=-1  

\noindent\textbf{Cost volume pyramid.}
We provide a finer analysis of the three scales in the cost-volume pyramid in Table~\ref{tab:2}. 
The finest-scale volume preserves local matching details but provides less contextual support for ambiguous regions, resulting in a 3-pixel error of 2.03 on KITTI 2015 and an EPE of 1.30 on Scene Flow. 
Adding the middle-scale cost volume reduces the errors to 1.91 and 1.22, respectively, indicating that intermediate contextual aggregation helps resolve local matching ambiguity. 
Further incorporating the coarsest-scale cost volume improves the results to 1.81 and 1.08, showing that coarse global geometry is beneficial for textureless regions, repetitive patterns, and large-disparity structures. 
These results demonstrate that different scales provide complementary geometric cues: fine scales retain local disparity details, while coarser scales enlarge the receptive field and stabilize disparity estimation. 
Therefore, the cost-volume pyramid supplies a stronger multi-scale initial geometry prior, which is further refined by the subsequent saliency-conditioned diffusion module.

\noindent \textbf{Effects of Saliency Attention Perception.} With the cost-volume pyramid and four-step diffusion fixed, adding the complete SAP module reduces the KITTI 2015 3-pixel error from 1.74\% to 1.52\% and the Scene Flow EPE from 0.90 to 0.51. Removing SSB while retaining CSB increases these errors from 1.52\% and 0.51 to 1.63\% and 0.76, respectively. Removing CSB while retaining SSB increases them to 1.69\% and 0.81. These controlled results show that the spatial- and channel-wise blocks provide complementary information for disparity refinement.

\noindent\textbf{Comparison with other attention mechanisms.}
To further evaluate SAP, we compare it with representative attention mechanisms under the same framework, including SE~\cite{hu2018squeeze}, CBAM~\cite{woo2018cbam}, GC block~\cite{cao2019gcnet}, and Non-local block~\cite{wang2018non}. We replace only the attention module while keeping the cost-volume pyramid, diffusion refinement, training protocol, and loss function unchanged. As shown in Table~\ref{tab:attention_comparison}, every evaluated attention variant reduces both errors relative to the baseline without attention. \re{SAP yields the lowest errors among these variants, with a 1.52\% 3-pixel error on KITTI 2015 and 0.51 EPE on Scene Flow.} Unlike generic channel or spatial recalibration and global-context aggregation, SAP explicitly models spatial- and channel-wise saliency to emphasize high-frequency geometric cues used by the subsequent diffusion refinement.

\begin{table}[!t]
  \centering
  \caption{Quantitative comparison of attention mechanisms on the KITTI 2015 validation split and Scene Flow test set. The baseline excludes attention but retains the multi-scale cost volume and four-step diffusion refinement.}
  \label{tab:attention_comparison}
  \renewcommand\tabcolsep{11.0pt}
  \resizebox{0.99\linewidth}{!}{
  \begin{tabular}{lcc}
    \toprule
    {Attention Module} & {KITTI 2015 \re{3-pixel error (\%)}} & {Scene Flow EPE} \\
    \midrule
    Baseline (None) & 1.74 & 0.90 \\
    SE~\cite{hu2018squeeze} & 1.70 & 0.85 \\
    CBAM~\cite{woo2018cbam} & 1.67 & 0.80 \\
    GC block~\cite{cao2019gcnet} & 1.65 & 0.76 \\
    Non-local~\cite{wang2018non} & 1.61 & 0.70 \\
    \midrule
    \rowcolor{gray!20}\textbf{\re{SAP (ours)}} & \textbf{1.52} & \textbf{0.51} \\
    \bottomrule
  \end{tabular}
  }
\end{table}

\noindent\textbf{Comparison of refinement strategies.}
\re{We compare refinement operators under a controlled stereo matching framework. All variants share the same cost-volume pyramid and initial disparity estimator. Unless the condition itself is ablated, CNN, ConvGRU, and diffusion receive the same initial disparity map $\mathbf{D}_{\mathrm{ini}}$ and confidence-guided SAP feature $\mathbf{C}'$. The CNN applies one feed-forward residual update, whereas ConvGRU and diffusion each perform four updates. The diffusion-without-SAP variant intentionally receives only $\mathbf{D}_{\mathrm{ini}}$. All variants are trained end-to-end with the same Scene Flow and KITTI splits, $256\times512$ training crops, maximum disparity of 192, batch size of 6, Adam optimizer, learning-rate schedule, loss functions, and multi-scale loss weights. Therefore, the refinement operator and the explicitly indicated condition ablation are the controlled differences among the settings. As shown in Table~\ref{tab:refinement_comparison}, the SAP-conditioned diffusion variant yields the lowest errors in this comparison, with a 1.52\% 3-pixel error on KITTI 2015 and an EPE of 0.51 on Scene Flow.}

\begin{table*}[!t]
\rcap
\centering
\caption{
Controlled comparison of refinement strategies on the KITTI 2015 validation split and Scene Flow test set. $\mathbf{D}_{\mathrm{ini}}+\mathbf{C}'$ denotes concatenation of the initial disparity map and confidence-guided SAP feature. All variants use the same stereo matching backbone and training protocol; the listed refinement operator and condition are the controlled differences.
}
\label{tab:refinement_comparison}
\renewcommand\tabcolsep{12.0pt}
\resizebox{0.90\textwidth}{!}{
\begin{tabular}{lcccc}
\toprule
\makecell{Refinement\\Strategy}
& \re{Condition}
& \re{Updates}
& \makecell{KITTI 2015\\\re{3-pixel error (\%)} $\downarrow$}
& \makecell{Scene Flow\\EPE $\downarrow$} \\
\midrule
\re{Initial disparity without refinement}
& \re{--} & \re{0} & 1.95 & 1.25 \\
CNN refinement
& \re{$\mathbf{D}_{\mathrm{ini}}+\mathbf{C}'$} & \re{1} & 1.82 & 1.02 \\
ConvGRU refinement
& \re{$\mathbf{D}_{\mathrm{ini}}+\mathbf{C}'$} & \re{4} & 1.73 & 0.81 \\
\re{Diffusion refinement without SAP}
& \re{$\mathbf{D}_{\mathrm{ini}}$} & \re{4} & 1.74 & 0.90 \\
\midrule
\rowcolor{gray!20}
\re{SAP-conditioned diffusion refinement}
& \re{$\mathbf{D}_{\mathrm{ini}}+\mathbf{C}'$} & \re{4} & \textbf{1.52} & \textbf{0.51} \\
\bottomrule
\end{tabular}
}
\end{table*}

\re{\noindent\textbf{Effects of iterative diffusion.}
Compared with the four-update ConvGRU variant, the four-step diffusion variant lowers the KITTI 2015 3-pixel error by 0.21 percentage points and the Scene Flow EPE by 0.30 pixels under the shared condition and training protocol.}

\begin{table}[!t]
\begin{center}
\rcap
\renewcommand\tabcolsep{15.0pt}
\caption{Runtime per stereo pair and estimation accuracy for different numbers of reverse diffusion steps. Runtime is measured on an NVIDIA RTX 4090 under the same inference setup. The 3-pixel error is evaluated on the KITTI 2015 validation split, and EPE on the Scene Flow test set. Four steps are used by default as an accuracy--efficiency compromise.}
\resizebox{0.99\linewidth}{!}{
\begin{tabular}{c|ccc}
\toprule
 \re{Diffusion steps} & \re{\makecell{KITTI 2015\\3-pixel error (\%)}} & \re{\makecell{Scene Flow\\EPE}} & \re{Runtime (s)}  \\  \midrule
 
1 &	1.82 &1.03	 &	0.15      \\
2 & 1.64 & 0.74	& 0.19	    \\
\rowcolor{Gray} 4 &1.52  &0.51 &0.28     \\ 
8 & 1.50 & 0.48 & 0.45   \\ 
12 &	1.49 & 0.47 & 0.61    \\ 
  \bottomrule
\end{tabular}
}
\label{efficiency}
\end{center}
\end{table}

\re{\noindent\textbf{Efficiency Analysis.}
The CNN variant requires one refinement-module evaluation, whereas ConvGRU and diffusion require four sequential updates. Their exact relative latency depends on the module implementation; separate CNN and ConvGRU timings were not used to make an efficiency claim. Under the NVIDIA RTX 4090 timing setup, four-step StereoDiffuer requires 0.28~s per stereo pair. Diffusion is selected for its measured accuracy and boundary-refinement results rather than for computational superiority.}

\re{Table~\ref{efficiency} further reports the accuracy--runtime trade-off as the number of reverse diffusion steps changes. The 3-pixel error is computed on the KITTI 2015 validation split, and EPE is computed on the Scene Flow test set. Increasing the number of steps from four to eight reduces the 3-pixel error from 1.52\% to 1.50\% and the EPE from 0.51 to 0.48, while increasing runtime from 0.28~s to 0.45~s. Twelve steps further increase runtime to 0.61~s but yield only small additional changes. We therefore use four reverse steps as the default accuracy--efficiency compromise.\looseness=-1}

\subsection{Discussion on Generalization and Robustness}

\noindent\textbf{Generalization to other domains.}
Although our experiments focus on Scene Flow and KITTI, the proposed framework is not tied to a specific semantic category or driving-specific prior. 
StereoDiffuer refines disparity estimation using generic stereo geometry cues, including the initial disparity map, cost-volume confidence, and saliency-aware geometric details. 
Therefore, SAP-conditioned diffusion refinement could be adapted to other stereo domains, such as indoor robotics, AR/VR, industrial inspection, and remote sensing, where boundary preservation and reliable geometric reconstruction are also important. 
Nevertheless, different domains may introduce substantial distribution shifts in camera baseline, disparity range, texture statistics, illumination, sensor noise, blur, and occlusion patterns. 
For such scenarios, several extensions can be considered, including target-domain fine-tuning, disparity-range calibration according to camera geometry, self-supervised adaptation with photometric and left-right consistency constraints, and degradation-aware saliency estimation. 
These strategies may improve the robustness of StereoDiffuer when transferring to domains beyond the evaluated benchmarks.

\noindent\textbf{Discussion on blur robustness.}
Motion blur is a common degradation in real-world stereo systems, especially in dynamic driving scenarios with fast camera or object motion. 
Since blur suppresses high-frequency image structures, it may weaken edge responses, object boundaries, and thin structures that are important for reliable cost-volume construction and saliency-guided disparity refinement. 
Recent deblurring studies provide useful insights for addressing this issue. 
MC-Blur~\cite{zhang2023mc} introduces a comprehensive multi-cause deblurring benchmark covering real-world and synthetic blur from different factors, such as motion and defocus, highlighting the diversity of blur degradations in practical scenarios. 
DBLRNet~\cite{zhang2018adversarial} models spatio-temporal information with 3D convolutions and adversarial learning to recover sharp details from blurred videos. 
Inspired by these works, several extensions could improve the robustness of our framework under blurred inputs. 
First, blur-aware augmentation can be incorporated during training to expose the cost-volume pyramid and SAP module to different motion-blur kernels and blur intensities. 
Second, a degradation-aware saliency estimator could be introduced to distinguish reliable geometric structures from blur-induced weak responses, so that diffusion refinement does not rely excessively on unreliable high-frequency cues. 
Third, the confidence map used in collaborative condition construction can be further modulated by blur uncertainty, reducing the contribution of severely blurred regions and encouraging stronger contextual support from neighboring reliable areas. 
Finally, for video stereo settings, an optional deblurring or spatio-temporal restoration branch, following the spirit of DBLRNet, could be integrated before disparity refinement to recover sharper features for matching. 
We leave a full blur-robust stereo matching extension as future work.


\section{Conclusion}
In this work, we present StereoDiffuer, an iterative diffusion-based stereo matching framework that explicitly models geometric details and progressively refines disparity estimates. SAP extracts spatial- and channel-wise saliency cues, while cost-volume confidence selects reliable geometric information to condition the reverse diffusion process together with the initial disparity map. \re{Experiments on Scene Flow and KITTI show that StereoDiffuer achieves competitive overall accuracy and provides improvements in the evaluated boundary and foreground regions. These results support the usefulness of saliency-conditioned progressive refinement without implying universal superiority over existing methods.} We hope StereoDiffuer encourages further research on explicit geometric modeling and iterative refinement for stereo matching and related 3D-vision applications.

\bibliographystyle{elsarticle-num-names} 
\bibliography{\assetpath ref,\assetpath selfcite}

\end{document}